\documentclass{article}
\usepackage{latexsym,amsmath,amssymb,amsfonts,graphicx}
\usepackage{amsmath}
\usepackage{amssymb}
\usepackage{float}
\usepackage{scalefnt}
\usepackage{ifthen}
\usepackage{url}
\usepackage{graphicx}
\usepackage{colortbl}
\usepackage{multirow}
\usepackage{array}

\newcommand{\MI}{{\it MI}}

\title{Meta-learning for feature selection}
\author{Ben Goertzel, Nil Geisweiller and Chris Poulin}

\begin{document}
\maketitle

\begin{abstract}
A general formulation of optimization problems in which various candidate solutions may use different feature-sets is presented, encompassing supervised classification, automated program learning and other cases.   A novel characterization of the concept of a ``good quality feature'' for such an optimization problem is provided; and a proposal regarding the integration of quality based feature selection into metalearning is suggested, wherein the quality of a feature for a problem is estimated using knowledge about related features in the context of related problems.   Results are presented regarding extensive testing of this "feature metalearning" approach on supervised text classification problems; it is demonstrated that, in this context, feature metalearning can provide significant and sometimes dramatic speedup over standard feature selection heuristics.
\end{abstract}

\section{Introduction}

The critical importance of feature selection in classification and other optimization tasks is well understood by all real-world practitioners of these technologies, yet is relatively under-addressed in the theoretical literature.  Some rigorous characterizations of feature quality in terms of concepts like mutual information and separability have been articulated (see \cite{Blachnik}, \cite{Duch}, \cite{Kachel} for reviews) but these do not yet constitute a comprehensive theory of feature quality.  

Here we present a novel theory of feature quality in a very general setting.  We begin by describing a general class of optimization problems in which various candidate solutions may use different feature-sets.  This class encompasses supervised classification, automated program learning and other cases.   We then present a formal characterization of what constitutes a ``good quality feature'' or a ``good quality feature subset'' for such an optimization problem.  While new in many respects, the core ideas come from previous practical work in feature quality estimation in the context of supervised classification of microarray and SNP data \cite{Biomind}.

Finally we discuss the application of these ideas to metalearning \cite{Brazdil}.   Our definition of feature quality could be used straightforwardly to drive feature selection, but the issue that arises here is that a fairly large knowledge base of potential problem solutions may be needed in order to get a reliable estimate of feature quality.  We suggest that metalearning may provide a way around this issue in many cases, via allowing knowledge about the impact of  features on solution quality to be transferred from one problem to other related problems.   That is, the quality of a feature for a problem may be estimated using knowledge about related features in the context of related problems.  In some contexts, this may be a more important variety of metalearning than the more common application of metalearning to algorithm selection or parameter tuning.   Sometimes feature selection may actually be the hard part of a machine learning problem, and the place where the assistance of metalearning is most badly needed.

\subsection{Evaluation of Feature Metalearning for Text Classification}

The bulk of the paper describes some computational experiments carried out
in order to provide an initial exploration of the proposed
approach to feature metalearning.  These experiments show that feature
metalearning has significant capability to speed up text classification performance.

In these experiments, we created a database
called the MetaDB, which stores learning problems together with the data-features found most useful
for solving them; and then to use the information in the MetaDB to
initialize the feature selection process for new problems, thus accelerating the feature
selection process for these new problems.   This is a very general concept; here we
explore it in the specific context of binary text classification, but we believe the concept deserves
to be explored much more broadly.

Specifically, in the experiments reported here, we have applied
feature metalearning to techtc300 text classification dataset
collection \cite{techtc300, DGM04}, as well as similar collections
generated using software we developed specially for this experiment
\cite{techtcbuilder}.  The datasets in this collection are all built
upon the DMOZ database \cite{DMOZ} that classifies a large number of
websites according to a certain ontology. Each dataset contains texts
of websites that belong to a certain category vs texts of websites
belonging to another category. An example dataset in the collection
would be:

\begin{itemize}
\item 278 documents (i.e. websites) belonging to the
  positive category $$Top/Arts/Music/Instruments/$$
\item 257 documents belonging to the negative category
  $$Top/Health/Medicine/Education$$
\end{itemize}

\noindent The  techtc300 collection contains 300 different
datasets of this nature.  

In the experiments we report here, each such dataset is pre-processed
and transformed into feature vectors where each feature marks the
presence or absence of a word in its corresponding document, and the
target associated with each feature vector indicates whether the
document belongs to the positive category or not. This is a simplistic
approach to assigning feature vectors to texts, but proved adequate
for initial exploration of the feature metalearning concept.

Crawling the DMOZ database allows one  to generate very large
collections of datasets.  However the datasets within a given collection can be
rather distant semantically, which increases the difficulty of enabling
effective feature metalearning learning.  However, we managed nevertheless to
achieve positive feature metalearning results.

Specifically, we have shown that, in the text classification context at least,

\begin{itemize}
\item If the collection of datasets is sufficiently "dense", in the sense that 
a dataset generally has some fairly close neighbors in dataset-space, then
feature metalearning gives positive speedup (over a more straightforward,
non metalearning based feature selection process) and doesn't degrade the quality
of the final answer
\item The speedup is greatest in the case where one needs a quick answer
and doesn't allocate enough time to the problem-solving process to find an
optimal feature set, but only needs a "best I can find in the time available"
feature set
\end{itemize}

\noindent At the end of the report we will present some of our ideas for how
to surmount or weaken these limitations, which we did not have time to explore yet.

We believe the feature metalearning approach has great promise to dramatically
accelerate machine learning in cases where there is a large number of relatively similar problems
with features drawn from a common feature space.  The initial algorithms reported here
have sufficed to demonstrate the viability of the approach, and explore some of its properties.
Refining them via future research should ultimately lead to the development of extremely
powerful feature metalearning systems.

\section{A Formal Characterization of Feature Quality}
\label{sec:formal}

\subsection{A General Class of Optimization Problems}

We begin by articulating a fairly general class of optimization problems to which the problems of ``feature selection'' and ``feature quality'' pertain.

Suppose we have a maximization problem with objective function
$\phi:G\rightarrow W$, where $W$ is a positive bounded subset of the
real line, and $G$ is a space of functions so that for some fixed sets
$B$ and $\theta = \prod_{i \in I} \theta_i$ (the Cartesian product of
all ``variables'' or ``features''),  each $g \in G$ has the form: $g: A \rightarrow B$ for some $A =  A_g = \prod_{i \in J} \theta_i, J \subseteq I$.

For instance, one may have $B$ = the real line, and $\theta$ = 100-dimensional real space.  Then, each $g \in G$ would map some n-dimensional space for $n \leq 100$ into some bounded subset of the real line.  The maximization problem then involves finding some $g$ of this nature, that maximizes some criterion.

Or more interestingly, one may have a traditional categorization problem, in which case $B$ is a set of category labels, and each $g$ is a ``classification model'' assigning a list of feature values to a category.  In this case the objective function $\phi$ calculates the F-measure or some other measure of the quality of the classification model. 

Or one may have an automated program learning problem, in which the $g$ are programs mapping sets of input variables into outputs lying in $B$.

In any of these situations, one pertinent question is: which ``features'' $\theta_i$ are most useful for solving the maximization problem?  And which feature-sets are most valuable?  This is an algorithmic question (how to find the useful features) but also a conceptual question (how to define what it means for a feature to be useful).  

\subsection{Formalizing Feature Quality}
 
Next, in order to define the quality of an individual feature $\theta_i$ with respect to an objective function $\phi$, we will assume that we have some measure $q(\theta_i,g)$ of the quality of a feature $\theta_i$ with respect to a particular function $g \in G$.   We would like $q(\theta_i,g) \in [0,1]$, where for instance

\begin{itemize}
\item if the variable $\theta_i$ is not used at all in $g$ then $q(\theta_i,g) =0$
\item if changing the value of $\theta_i$ changes the output value of $g$ (at least sometimes), then $q(\theta_i,g) > 0$
\end{itemize}

\noindent Our definition of feature quality for $\phi$ can be used with  many different approaches to $q$; but a little later we will describe one particular approach to $q$ that seems potentially promising.

Given the above, we define the quality of feature $\theta_i$ with respect to objective function $\phi$ as

%
%
%

$$
Q(\theta_i, \phi) = \int q(\theta_i, g) T(\phi(g)) d\mu(g)
$$

\noindent where $\mu(g)$ represents a prior distribution over the $g$
functions, e.g. perhaps a simplicity-based prior such as the
Solomonoff prior. $T$ represents a fitness distortion.

This tells, basically, the average degree to which variation in the
feature $\theta_i$ affects the output of the functions $g$, with a
weighting in the average to value $g$ that are close to optimizing
$\phi$.  It is a generalization of the definition of ``feature
importance'' used in the OpenBiomind toolkit
\url{http://code.google.com/p/openbiomind/}, used to good effect in a
number of practical genomics classification problems, and described in
\cite{Biomind}.

We may start with $T(w) = \left( \frac{\phi(g)}{sup_g \phi(g)} \right)
^p$, that is the fitness distortion normalizes the fitness and
emphasizes good fits. The exponent $p$ weights how much the
near-optimality of $g$ figures into the calculation.  Of course other
distortion functions besides the $p$-power could be used, here, but
for the moment the $p$-power seems to give adequate flexibility.

At first glance it may seem that this feature quality measure looks at each variable in isolation, but actually it considers interactions, because the functions $g$ combine the variables in various ways.

The next question is how to consider multiple variables jointly, e.g. how to think about the combined utility of two features $\theta_i$ and $\theta_j$ (or sets of three or more features).  The simplest approach is just to consider a set $S$ of features as if it were a single feature, and apply the above definition directly to this single amalgamated feature, i.e.

$$
Q(S, \phi) = \int q(S, g) T(\phi(g)) d\mu(g)
$$


\noindent where $S$ is a set of more than one $\theta_j$. This works
perfectly well so long as one has a definition of $q(S, g)$.

Finally, note that if $T(\phi(g))$ is defined as the probability of
function $g$ meeting some objective $o$, then $Q(S, \phi)$ is the
expectation of the quality of $S$ (as defined by $q$) relative to the
given optimization problem.

\subsection{Feature Quality Relative to a Specific Function $g$}

One potentially interesting way to define $q$ is using the Fisher information.  To apply the Fisher information here, first we will translate the function $g$ being optimized into a probability density function.  So we can consider the output value of $g$ as a random variable $X_{g}$, and the density $f_{g}(X_{g})$ is then defined by the standard equation $ P (X_{g} \in D) = \int_{x \in D} df_g(x) $.

At each particular $\theta$ vector, we may then define the Fisher information $\mathcal{I}_g(\theta)_i$ corresponding to the variable $\theta_i$.   This tells, roughly speaking, how much the output of $g$ varies, as one varies $\theta_i$ in the neighborhood of $\theta$.  We  may then define

$$
q(\theta_i,g) = \frac { \int \mathcal{I}_g(\theta)_i d\nu(\theta)} { max_j \int \mathcal{I}_g(\theta)_j d\nu(\theta) }
$$

\noindent where $\nu$ is a prior distribution over $\theta$ values.

The same definition works for a feature-set S, i.e.

$$
q(S,g) = \frac { \int \mathcal{I}_g(\theta)_S d\nu(\theta) }{ max_{S'} \int \mathcal{I}_g(\theta)_{S'} d\nu(\theta) }
$$

\noindent (where $S'$ is assumed to be a feature-set drawn from $\theta$ with the same cardinality as $S$), the only complication here being that one must define $\mathcal{I}_g(\theta)_S$ more subtly, e.g. as the average Fisher information $\mathcal{I}_g(\theta)$ calculated along all geodesic paths from $\theta$ to other vectors $\theta'$ obtainable via varying only features in S.

\section{Methodology}
In this section we explain in detail the steps of the methodology
employed in our experiments in feature metalearning for text classification.

\subsection{Overview of the Meta-algorithm}
\label{sec:metaalgo}
Given a new problem D (i.e. a classification task taken from some
\emph{techtcX} collection, where X typically represents the size of
the collection), the chain of tasks follows:
\begin{enumerate}
\item Get initial features from the \emph{MetaDB} (explained below),
  supposedly belonging to the set of best features of D, by selecting
  the best features of problems nearby D (see Section
  \ref{sec:feattrans}).
\item Run a feature-search method for feature selection using the set of features obtained in the
  previous step to initialize the search and hopefully speed it up
 (see Section \ref{sec:featalgo}).
\item Run the learning task with the set of features selected in the
  previous step (see Section \ref{sec:leartask}).
\item Given the information obtained in the previous step, update the
  MetaDB; that is, associate the problem with its best features plus
  other useful information such as their quality (see Section
  \ref{sec:upmdb}).
\item Wait for a new problem and go back to step 1.
\end{enumerate}

\subsection{MetaDB}
The MetaDB is a repository associating problems already explored to
features found to be useful to that problem. Technically speaking it
is an XML file containing for each problem
\begin{enumerate}
\item A unique ID of the problem (here, the path of its corresponding
  dataset).
\item The set of features with their feature qualities, $q$ and $Q$ as
  defined in Sections \ref{sec:fitfunc} and \ref{sec:upmdb}
  respectively.
\item Information such as the minimum and maximum score obtainable for
  that dataset to normalize the fitness function of the learning
  problem.
\item The set of candidates with their scores obtained during
  learning.
\end{enumerate}

Additionally all problems already explored are stored in a covertree
\cite{Beygelzimer06covertrees} for fast retrieval.

\subsection{Feature Transfer}
\label{sec:feattrans}
In absence of metalearning the search starts from the empty set. With
feature transfer the search starts from the feature set given during
the step 1 of the meta-algorithm described in Section
\ref{sec:metaalgo}.

Positive (resp. negative) transfer is said to have occurred if the
search is faster (resp. slower) when starting from that given feature
set rather than from the empty set, with similar feature quality
obtained in both cases.

\subsubsection{Formula to select features to transfer}

Seeding the feature search for a new problem via selecting the entire feature set of nearby problems, hasn't proved effective. This strategy might work if the problems involved was
composed of very similar problems (for instance if all datasets were
restricted to sentiment analysis of similar types of text). But in the situation
we explored in our experiments,  each dataset corresponds to a rather different classifier, sometimes pretty distant from each other
semantically, such as Music vs Health, Business vs Movie, etc. So
instead we chose to consider each feature of a new problem separately and decide
based on appropriate heuristics whether it may be a good feature for the new
problem or not.

The heuristics utilized may be interpreted as means of approximation

$$P_D(s) = {\it probability\ that\ feature\ s\ is\ in\ the\ best\ feature\ set\
  of\ D}$$

\noindent where $D$ is the dataset corresponding to the new problem
instance. Then given a {\it transfer threshold} $t$ one selects all
features $s$ belonging to the best feature sets of the k-nearest
datasets so that $P_D(s) \ge t$. By setting the value $t$ adequately
we hope to filter feature transfer so that positive ones occur
significantly more often than negative ones.

\subsubsection{Our heuristic}
\label{sec:heu}

Estimating $P_D$ is not trivial, and in our work so far we have utilized
a simple heuristic (which no doubt could be improved substantially):

$$P_D(s) = \frac{\sum_{i=1}^k w(d(D, D_i)) \times Q(s, D_i)}{\sum_{i=1}^k
  w(d(D, D_i))}$$

\noindent where
\begin{itemize}
\item $D_1$, ..., $D_k$ are the k-nearest neighbors (according to the
  Jensen-Shannon divergence, briefly explained in Section
  \ref{sec:jenshan}),
\item $d(D, D_i)$ is the square root of the Jensen-Shannon divergence,
\item $w(d) = 1/(c+d^2)$, where $c$ is a positive constant
  (arbitrarily fixed to 0.001 in our experiments),
\item $Q(s, D_i)$ is the quality of feature $s$ of the problem
  corresponding to dataset $D_i$.
\end{itemize}

\noindent The formula expresses that: if a feature is important w.r.t. nearby
datasets of $D$, then it is likely to be important w.r.t. $D$, and
this likelihood increases with the importance of the feature and the
closeness of the neighbors.   The definition of $w(d)$ is somewhat ad hoc, though functional, and
will likely be refined via further experimentation.

\subsubsection{Problem Distance: Jensen-Shannon Divergence}
\label{sec:jenshan}
In our practical work so far, we have used the square root of the Jensen-Shannon divergence to measure
distance between datasets.  We made this choice because of its simplicity and manageability -- it is 
always within $[0, 1]$, and it is a true metric, which is important so that
fast KNN query algorithms such as covertree can be used.

The Jensen-Shannon divergence (JSD for short) is defined as

$$\textit{JSD}(P, Q) = \frac{1}{2}\textit{KLD}(P, M) + \frac{1}{2}\textit{KLD}(Q, M)$$

\noindent where KLD is the Kullback-Leibler divergence and $M =
\frac{1}{2}(P+Q)$.

The following examples illustrate how this measure is used
here. Let's assume the following dataset $D_1$

$$
\begin{array}{|c|c|c|}
\hline
word1 & word2 & target \\
\hline
0 & 0 & 1 \\
0 & 1 & 1 \\
0 & 0 & 1 \\
1 & 0 & 0 \\
\hline
\end{array}
$$

\noindent where \emph{target} represents whether the document of a
certain row belongs to the positive ODP category corresponding to that
dataset. Similarly $D_2$

$$
\begin{array}{|c|c|c|}
\hline
word1 & word2 & target \\
\hline
1 & 1 & 1 \\
1 & 1 & 1 \\
1 & 0 & 1 \\
0 & 0 & 0 \\
\hline
\end{array}
$$

Let's compute the square root of the JSD between $D_1$ and $D_2$. We must
first represent the datasets as distributions, let's start with $D_1$:

$$P_{D_1}(\{target\}) = 2/4 = 0.5$$
$$P_{D_1}(\{word2, target\}) = 1/4 = 0.25$$
$$P_{D_1}(\{word1\}) = 1/4 = 0.25$$

\noindent then $D_2$:

$$P_{D_2}(\{word1, word2, target\}) = 2/4 = 0.5$$
$$P_{D_2}(\{word1, target\}) = 1/4 = 0.25$$
$$P_{D_2}(\emptyset) = 1/4 = 0.25$$

\noindent In this example $P_{D_1}$ and $P_{D_2}$ are entirely
disjoint; as a result

$$d(D_1, D_2) = 1$$

Let's consider the dataset $D_3$

$$
\begin{array}{|c|c|c|}
\hline
word1 & word2 & target \\
\hline
1 & 1 & 1 \\
0 & 1 & 1 \\
1 & 0 & 1 \\
0 & 0 & 0 \\
\hline
\end{array}
$$

\noindent with distributions

$$P_{D_3}(\{word1, word2, target\}) = 1/4 = 0.25$$
$$P_{D_3}(\{word2, target\}) = 1/4 = 0.25$$
$$P_{D_3}(\{word1, target\}) = 1/4 = 0.25$$
$$P_{D_3}(\emptyset) = 1/4 = 0.25$$

\noindent Some intersection exists between $P_{D_2}$ and $P_{D_3}$ and the
result of the square root of their JSD is

$$d(D_2, D_3) = 0.3945$$

As shown in Section \ref{sec:exp}, in our case the distance is often 1
and rarely goes below 0.9. That is because the datasets generated by
the techtc methodology contain many, many features, exceeding several
dozens of thousands across all datasets. The heuristic we are using
has been tuned in a way that accounts for this, and positive transfer
occurs nevertheless. However this poses a problem for the k-nearest
neighbor search as the dataset-space has a very large intrinsic
dimensionality. In our experiments, even when using the largest
collection, covertree based search would yield similar performance to
naive search.  There are certainly workarounds for this phenomenon,
but we did not have time to implement and evaluate them in our work so
far.  For instance, one could compute the JSD only over a projection
of generally important features, thus reducing the dimensionality of
the space.

\subsection{Feature Selection}
\label{sec:featalgo}
To select a good feature set we search the space of feature sets that
maximize a certain fitness function.
\subsubsection{Algorithm to search features sets}
The algorithm used to search the space of feature sets is a stochastic
variation of hillclimbing. Let $S$ be a set of features. Let $d$ be
the number of features to add or remove at once from $S$. Let $m$ be
the maximum number of evaluations.
\begin{enumerate}
\item Start with $S=\emptyset$, $d=1$ and $m = M$ (where $M$ is given
  by the user).
\item Randomly add and/or remove $d$ features from $S$, to generate
  feature sets $S_1$ to $S_h$ (so each $S_i$ is at edit distance $d$
  from $S$), where $$h=\min\left (\frac{m}{10}, |F|^d \right )$$ and $|F|$ is the
  total number of features in the dataset.
\item Set $m=m-h$.
\item Evaluate $S_1$ to $S_h$ (according to the fitness function
  defined below).
\item If there exists $S_i$ better than $S$, set $S = S_i$ and
  $d=1$. Otherwise increment $d$.
\item Go to step 2 unless $d = D$ or $m=0$.
\end{enumerate}
Here $D$ is fixed to 5.

\subsubsection{Fitness function to evaluate feature sets}
\label{sec:fitfunc}
We want feature sets which are informative but also as small as
possible for two reasons
\begin{enumerate}
\item The smaller the feature set, the faster the learning algorithm.
\item The larger the feature set the less accurate is its measure of
  information gain.
\end{enumerate}

With this in mind, the fitness function of a feature set is defined as

$$q(S) = \MI(S)^{2-r} \times c(S)^r$$

\noindent where $\MI(S) = H(S; Y)$ is the mutual information of $S$
and the output $Y$, and $c(S)$ is a function that measures the
confidence of $\MI(S)$. The parameter $r$ measures how important
confidence must be accounted for, the higher $r$ the more important
the confidence, the smaller the best feature set will be found.

By using such a fitness function we not only aim at informational
feature sets but also confident ones -- which should also minimize
overfitting during the learning process.
\subsubsection{Feature set size and confidence}
\label{sec:featconf}
Inspired by OpenCog's Probabilistic Logic Networks \cite{PLN}, we
could define $c(S)$ as the probability that $\MI(S)$ will be within a
certain interval $[L, U]$ after k more observations\footnote{a method
  sharing some similarities with statistical bootstrapping
}.

But those methods are rather costly and complex to implement so we
started with the following heuristic

$$c(S) = N/(N+b \times n)$$

\noindent where $n$ is the size of $S$ (i.e. the number of features)
and $b$ is a parameter.  More sophisticated and theoretically grounded
heuristics are possible and we have done some work in this direction,
but not implemented yet. That heuristic reflects the simple idea that
the larger the number of features the less confidence the estimation
of $\MI(S)$ will be.

Using this heuristic formula, if $b$ tends to 0, $c(S)$ tends to 1; for that reason we have considered
the parameter $b$ instead of the parameters $r$ defined in Section
\ref{sec:fitfunc} to modulate the importance of the confidence (that
way we only have one parameter to tweak).

\subsection{Learning Task}
\label{sec:leartask}
Once a set of features has been selected the dataset is filtered and
used as input of a classifier learner, here MOSES \cite{Looks2006},
\cite{Scalable}, a probabilistic evolutionary program learning
algorithm that we have used in many practical applications. The scores
output by MOSES correspond to
$$\textit{minus the number of classification errors}$$
as reported in Figures \ref{fig:msc1}, \ref{fig:msc2},
\ref{fig:msc3} and \ref{fig:msc4}.

\subsection{Update MetaDB}
\label{sec:upmdb}
The results output by MOSES are then used to update the MetaDB with
the feature quality measures. In Section \ref{sec:formal} the notion
of feature quality $Q$ is defined according to a fitness function to
keep the definition as general as possible.  In these specific
experiments with text classification, however, we work on a specific
class of fitness functions representing the error rate of a program
fitting a dataset $D$. In this context it is essentially equivalent to
just use $q(s)$ as defined in Section \ref{sec:fitfunc}. That is what
we have done in these experiments but it would certainly be
interesting to try with $Q$ as well.

\section{Experiments}
\label{sec:exp}
We have conducted several experiments on techtcX collections to
test the validity of those ideas.

\subsection{Method}
There are several ways to measure knowledge transfer, here we have
focused mainly on the speed to select features. For that purpose each
experiment are composed of the following steps

\begin{enumerate}
\item Run feature selection and learning over the entire dataset
  collection without any meta-learning taking place.
\item Mine the logs of the results obtained in the previous step and
  generate a list of \emph{accuracy targets}, containing the maximum
  score obtained by feature selection for each problem in the
  sequence.
\item Run feature selection and learning over the entire dataset
  collection according to the meta-learning algorithm in Section
  \ref{sec:metaalgo}, with the particularity that each feature
  selection process runs until it reaches its corresponding accuracy
  target defined in the previous step.
\end{enumerate}

In both cases (with or without metalearning) the feature selection stops
when a certain number of evaluations is reached.

Then we analyze the results and measure transfer learning in terms of
acceleration of feature selection during the metalearning phase (step 3) as
compared to the non-metalearning phase (step 1).

\subsection{Parameters}
The list of parameters we varied between our text classifications experiments is 
as follows:

\begin{itemize}
\item \emph{TCSize}: the size of the techtc collection. Here the
  possible sizes are 300, 500. The techtc300 was directly taken from
  \cite{techtc300}. The techtc500 was generated by our software
  \cite{techtcbuilder}. 
\item \emph{PUber, NUber}: the positive and negative uber categories
  where the subcategories are extracted to generate the datasets of the
  techtc collection. The techtc300 collection use both Top as uber
  category. The techtc500 uses Top/Arts/Music as uber positive
  category and Top/Science/Math as uber negative category.
\item \emph{MI}: the threshold of the mutual information used to
  pre-filter the techtc collection. The web is a rather messy place
  and it is not unusual to get dozens of thousands of features for a
  given dataset, lot of them being gibberish. For that reason we
  remove all features under a certain mutual information
  threshold. The higher the threshold the easier the feature selection
  task but with an additional risk of missing good features.
\item \emph{FE}: the number of evaluations allocated to feature selection (each
  evaluation measure how a feature set fits according to the fitness
  function defined in Section \ref{sec:fitfunc})
\item \emph{t}: the transfer threshold
\end{itemize}
Some other parameters have been left fixed
\begin{itemize}
\item The number of neighbors considered, $k=5$ in all experiments. We
  determined this number by conducting a few experiments which seemed
  to indicate that the number 5 works well.
\item The number of evaluations used during the learning tasks was
  also fixed to 10000.
\item The importance of the confidence in the feature selection
  fitness function (parameter $b$ defined in Section
  \ref{sec:featconf}) was fixed to $b=5$. This setting limits the number of features selected to around a
  dozen in average, thus speeding up the experiments.
\end{itemize}

\subsection{Results}

We have conducted hundreds of experiments, using the techtc300 and the
techtc500 collections. Following is a summary of some of the more
interesting results.

Most figures represent the speed-up with and without meta-learning as
a function of the iteration index of the sub-experiment in the
series. We expect that if significant meta-level learning occurs, the speed-up
will progressively increase as more problems have been
solved and their quality features recorded for future use. Specifically the speed-up is defined as

$$\text{Speed-up} = \frac{num\ evals\ without\ metalearning}
{num\ evals\ with\ metalearning}$$

We also report the arithmetic and geometric means of the speed-up, the
former is intuitively clear, but the latter is a better measure of the
overall speed-up.

\subsection{Experiments with the techtc300, using Top categories}
We present a series of experiments with the techtc300 collection,
about 300 classification problems generated from the top categories,
therefore possibly quite semantically distant.

\subsubsection{Low feature selection effort, low transfer threshold}
\begin{figure}
  \includegraphics[scale=0.38]{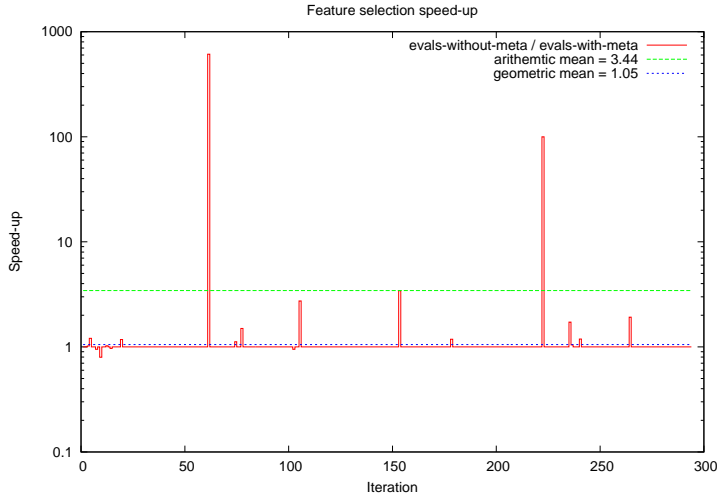}
  \caption{\label{fig:spup1}Speed-up for techtc300, $MI=0.001$, $FE=1000$,
    $t=0.09$}
\end{figure}

In this experiment, $MI=0.001$, $FE=1000$, $t=0.09$.
Figure \ref{fig:spup1} represents for each feature selection task the
speed-up (positive transfer) provided during metalearning as compared
without metalearning.

\begin{figure}
  \includegraphics[scale=0.38]{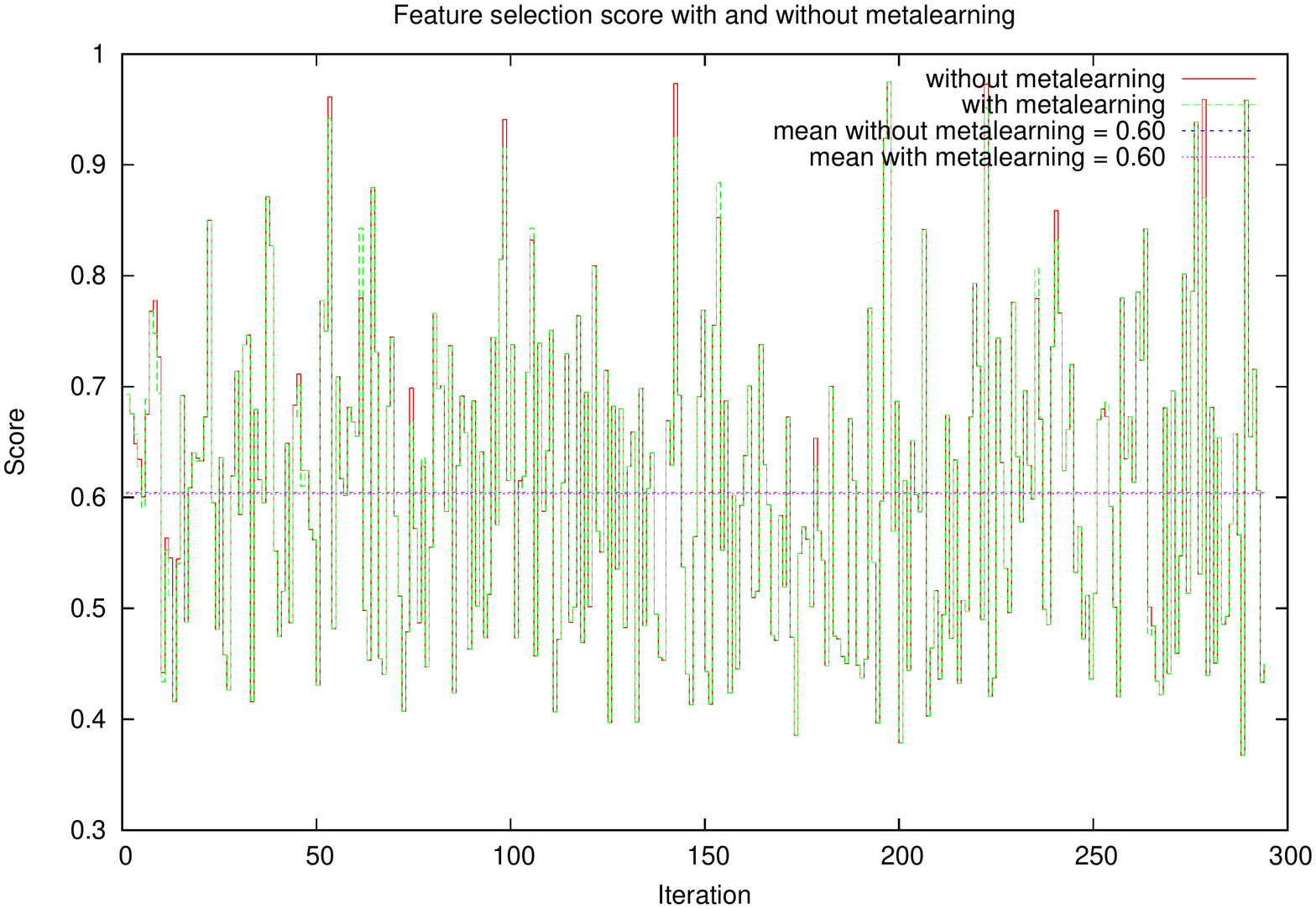}
  \caption{\label{fig:fsc1}Feature selection scores for techtc300,
    $MI=0.001$, $FE=1000$, $t=0.09$}
\end{figure}

The arithmetic mean of the speedup is 3.44, while the geometric mean
is 1.05.  The transfer threshold, 0.09, is quite low yet still yields
some positive transfer. The reason behind that, besides the fact that
the formula defined in Section \ref{sec:heu} is a heuristic rather
than the real probability, is that it does not account for the low
number of evaluations ($FE=1000$) used to reach the accuracy target
during the non-metalearning phase, as a consequence the accuracy
targets are rather low (as shown in Figure \ref{fig:fsc1}) and
therefore many more features can potentially speed-up the search.

\begin{figure}
  \includegraphics[scale=0.38]{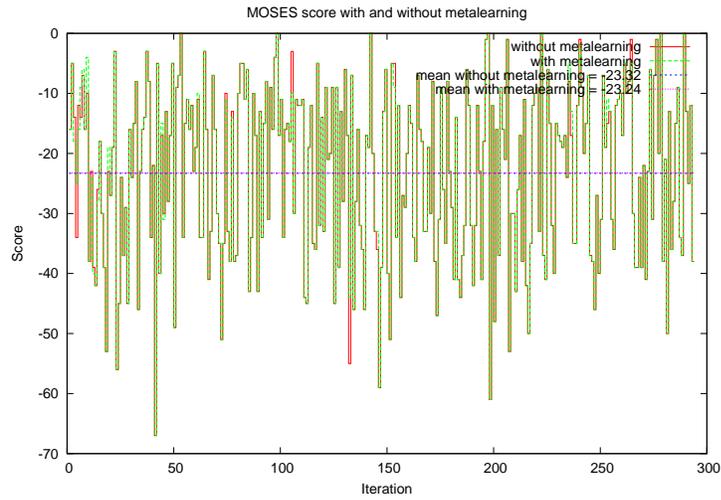}
  \caption{\label{fig:msc1}Learning scores for techtc300, $MI=0.001$,
    $FE=1000$, $t=0.09$}
\end{figure}

Finally, me can check that the speed-up is not at the expense of the
accuracy of the feature selection and the learning processes) as shown
in Figures \ref{fig:fsc1} and \ref{fig:msc1}.

\subsubsection{High feature selection effort, low transfer threshold}
\begin{figure}
  \includegraphics[scale=0.38]{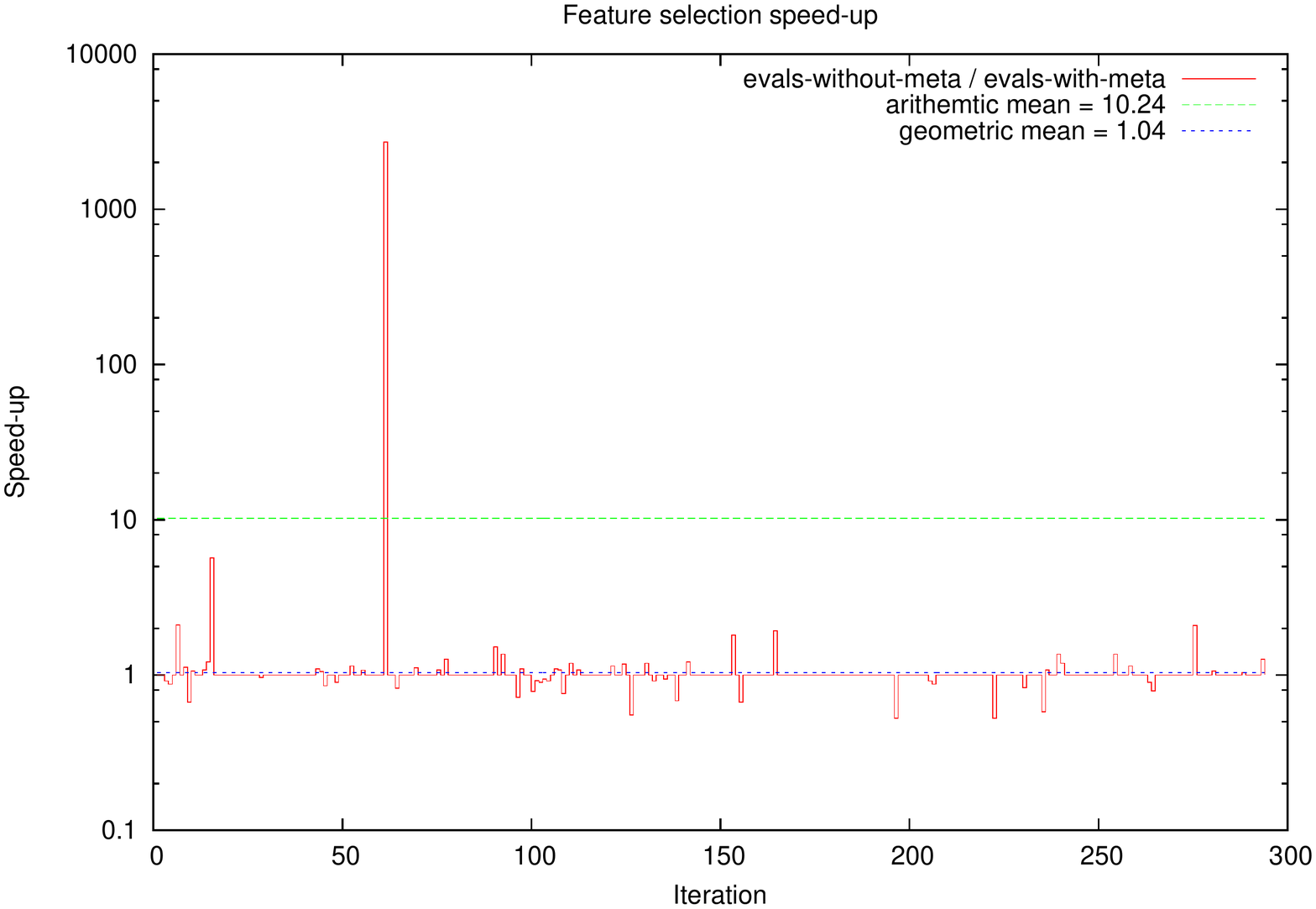}
  \caption{\label{fig:spup3}Speed-up for techtc300, $MI=0.001$,
    $FE=10000$, $t=0.09$}
\end{figure}

In this experiment we multiplied the feature selection effort by 10,
$FE=10000$, while maintaining the same transfer threshold, $t=0.09$.
As shown in Figure \ref{fig:spup3}, except for iteration 61, that led
to a massive speed-up, positive and negative transfers occurred
roughly equally often. After inspecting the log we could explain the
massive speed-up at iteration 61 by the fact that feature selection
without metalearning selected, after 2710 evaluations, \{submarin,
uss, repeat\} as best feature set, a rather small set with a high
score of 0.89. But during metalearning our heuristic selected an even
better feature set \{navi, radio, uss\} with a score of 0.9. As a
result there was only one evaluation during the feature selection with
metalearning as the accuracy target was immediately reached, yielding
the massive speed-up of 2710. This speed-up raises the arithmetic mean
up to 10.24, significantly higher than the one shown in Figure
\ref{fig:spup1} but the geometric mean is actually slightly lower
(1.04 instead of 1.05); as iteration 61 was an isolated case it does
not contribute enough to yield better overall performance.

\begin{figure}
  \includegraphics[scale=0.38]{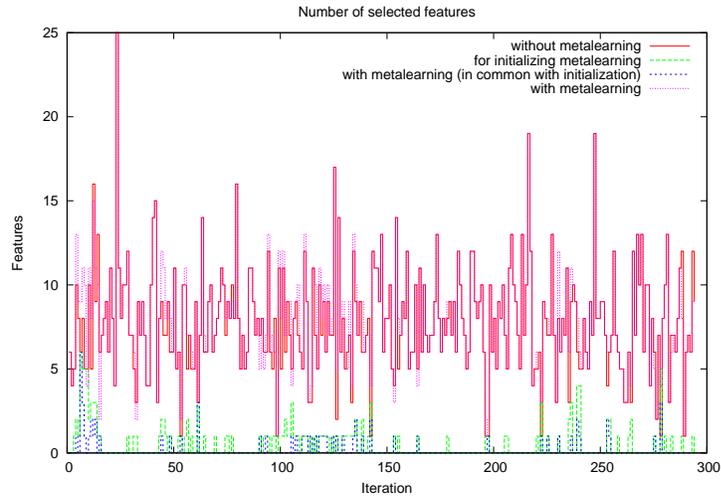}
  \caption{\label{fig:fn1}Number of features for techtc300, $MI=0.001$,
    $FE=10000$, $t=0.09$}
\end{figure}

Figure \ref{fig:fn1} shows the number of features selected without and
with metalearning as well as the number of feature transferred during
metalearning and the number of feature in common with the transferred
features and the selected feature during metalearning.

\subsubsection{High feature selection effort, very low transfer threshold}
\begin{figure}
  \includegraphics[scale=0.38]{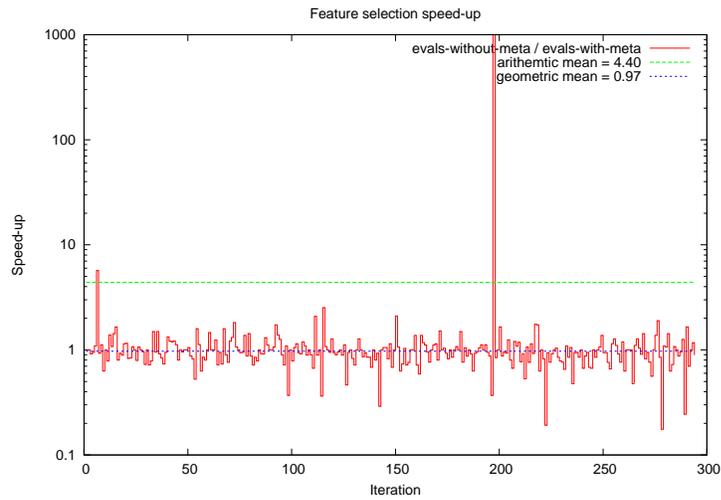}
  \caption{\label{fig:spup4}Speed-up for techtc300, $MI=0.001$,
    $FE=10000$, $t=0.01$}
\end{figure}

Decreasing the transfer threshold down to $t=0.01$ greatly increases
the number of transfers (see Figure \ref{fig:spup4}); however, as
above, except a happy accident raising the arithmetic speed-up to 4.4
in one case, the geometric mean is actually below 1, which means that
the overall performances have actually decreased during
metalearning. This shows the importance of setting the transfer
threshold high enough.

\subsubsection{High feature selection effort, very high transfer threshold}
\begin{figure}
  \includegraphics[scale=0.38]{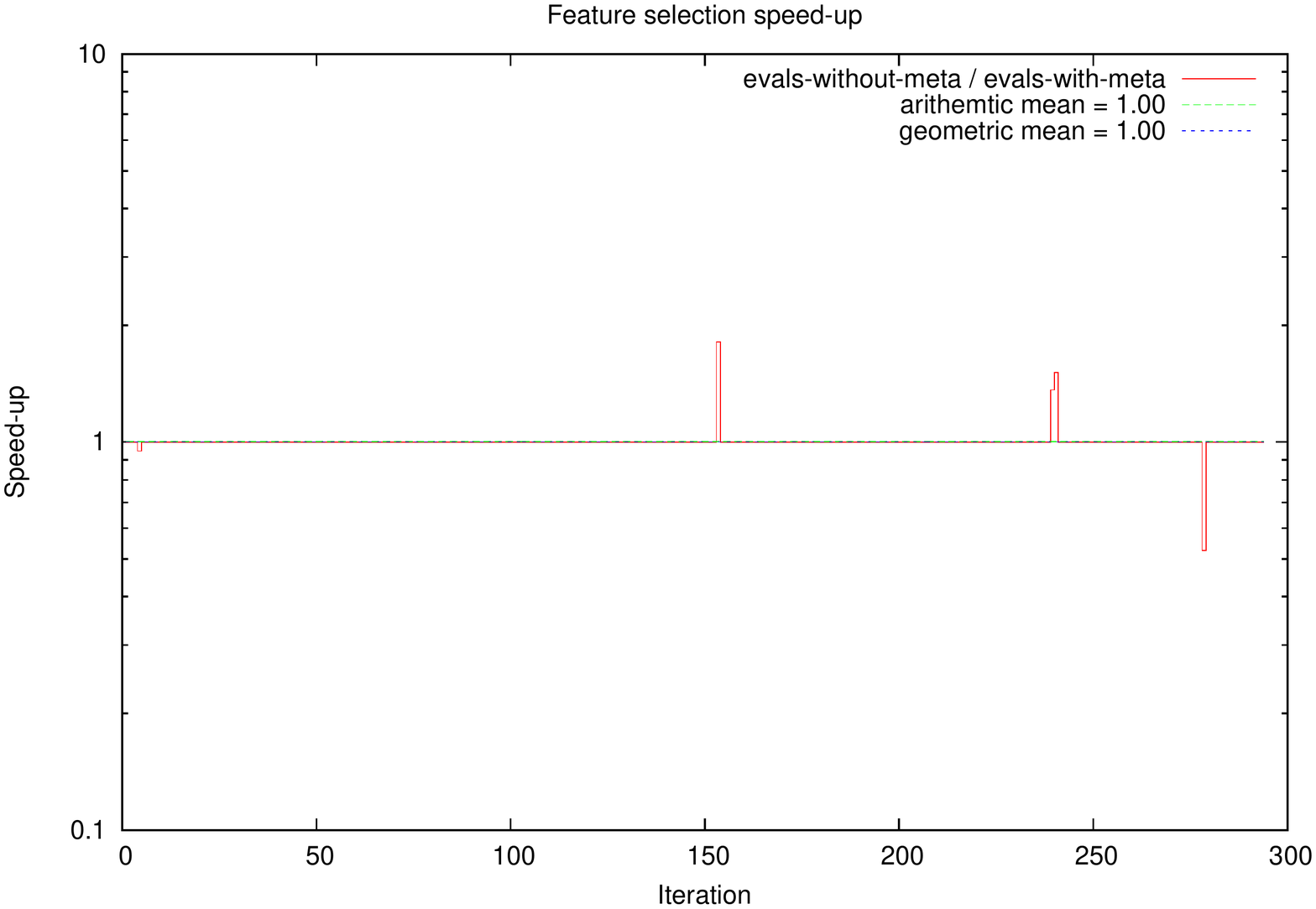}
  \caption{\label{fig:spup5}Speed-up for techtc300, $MI=0.001$,
    $FE=10000$, $t=0.2$}
\end{figure}

As shown in Figure \ref{fig:spup5}, increasing the transfer threshold
up to $t=0.2$ eliminates most transfer including the 2 very good ones
shown in Figures \ref{fig:spup3} and \ref{fig:spup4}.

\begin{figure}
  \includegraphics[scale=0.38]{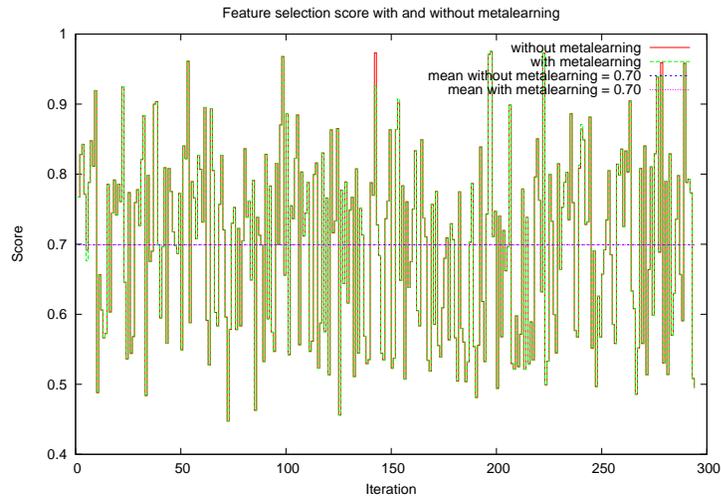}
  \caption{\label{fig:fsc2}Feature selection scores for techtc300,
    $MI=0.001$, $FE=10000$, $t=0.2$}
\end{figure}

\begin{figure}
  \includegraphics[scale=0.38]{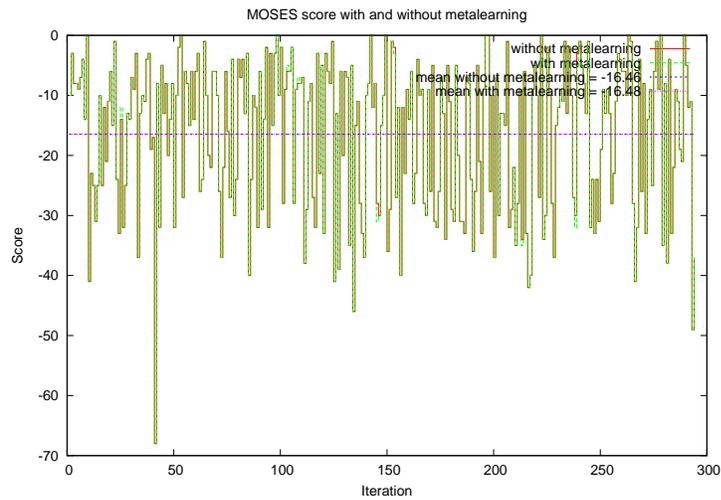}
  \caption{\label{fig:msc2}Learning scores for techtc300,
    $MI=0.001$, $FE=10000$, $t=0.2$}
\end{figure}

So as shown by this experiment and the previous one, increasing the
number of evaluations up to 10000 for the techtc300 does not allow us
to get any significant overall performance during
metalearning. However using 10000 evaluations instead of 1000 yields
significantly better scores with respect to feature selection as shown
in Figure \ref{fig:fsc2} (0.7 instead of 0.6 in Figure \ref{fig:fsc1})
and learning as shown in Figure \ref{fig:msc2} (-16.4 instead of -23.3
in Figure \ref{fig:msc1}). So transfer learning in the case of the
techtc300, although shown to happen and validate the idea, is not
really beneficial in practice. Add to that, if the overall speed-up
was more significant we would still need to take into account the
additional computational cost of transferring features -- which
consists mostly, as we will see in the next experiments, of the
calculations of the Jensen-Shannon divergences to find the nearest
datasets.

\subsection{Experiments with the techtc500, using Top categories
  Top/Arts/Music, Top/Science/Math}
That next series of experiments was done using the techtc500, generated
by us using \cite{techtcbuilder}. The results obtained below are not
directly comparable with the ones obtained with the techtc300
collection because
\begin{enumerate}
\item They have been generated with different software that probably
  crawl the web differently and filter the information differently.
\item The top-level categories for the techtc500 are much narrower
  (Top/Arts/Music for positive and Top/Science/Math for negative) than
  the ones of the techtc300 (Top for both positive and negative).
\item The number of samples for each dataset is higher than with the
  techtc300 (about 200 on average for the techtc300, and about 300 on
  average for the techtc500).
\end{enumerate}


\subsection{Low feature selection effort, high transfer threshold}
Here the pre-filtering threshold of the datasets is more aggressive
($MI=0.05$ instead of $MI=0.001$ as previously), mainly to decrease the
computational time of the experiments (so each experiment still takes
less than a day).

\begin{figure}
  \includegraphics[scale=0.38]{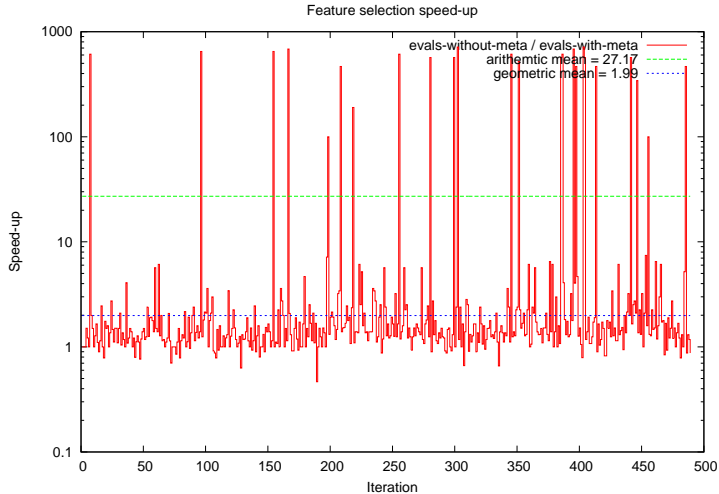}
  \caption{\label{fig:spup6}Speed-up for techtc500, $MI=0.05$,
    $FE=1000$, $t=0.1$}
\end{figure}

\begin{figure}
  \includegraphics[scale=0.38]{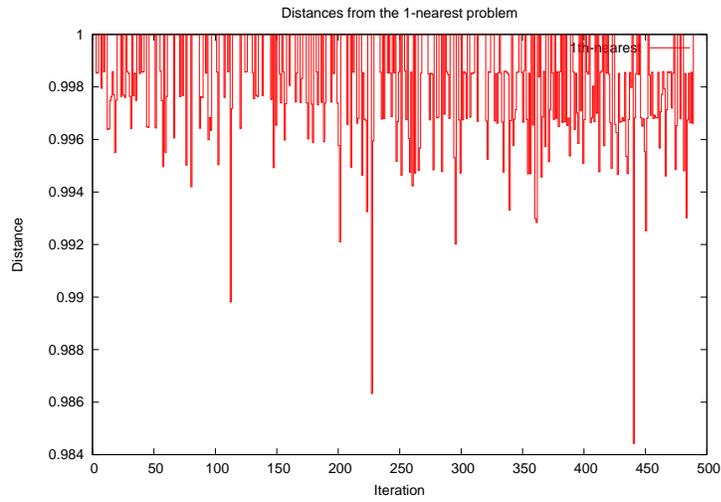}
  \caption{\label{fig:dk1}Distance to nearest neighbors for techtc500, $MI=0.05$,
    $FE=1000$, $t=0.1$}
\end{figure}

The number of evaluation for feature selection is rather low,
$FE=1000$. And the transfer threshold moderately high, $t=0.1$.  As
shown in Figure \ref{fig:spup6}, the results are quite encouraging,
with a speed-up arithmetic mean of 27.17 and a geometric mean of
1.99. Moreover there is a clear indication that the frequency of
massive speed-ups (occurring when the feature transfer find a better
feature set right away) increases. This is obviously due to the fact
that as measure as the experiment progresses the probability to find
nearby datasets increases, as corroborated by Figure \ref{fig:dk1}.

\begin{figure}
  \includegraphics[scale=0.38]{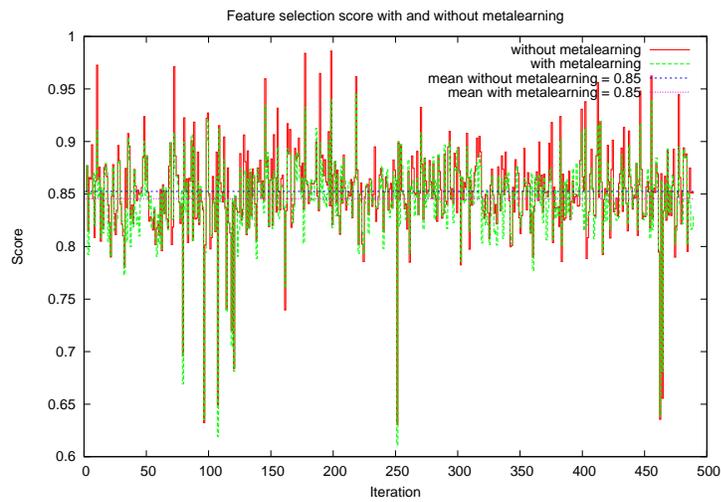}
  \caption{\label{fig:fsc3}Feature selection scores for techtc500,
    $MI=0.05$, $FE=1000$, $t=0.1$}
\end{figure}

\begin{figure}
  \includegraphics[scale=0.38]{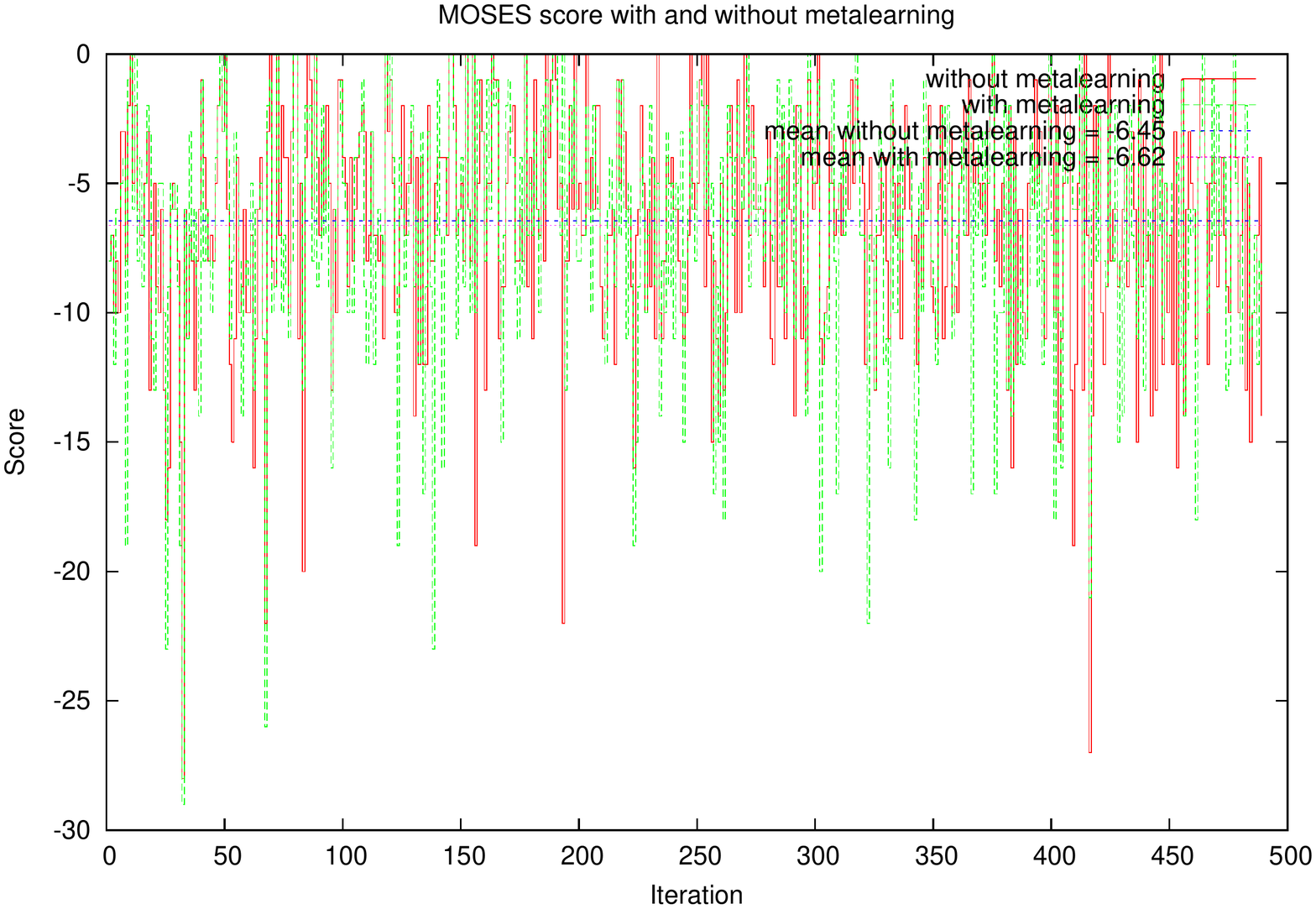}
  \caption{\label{fig:msc3}Learning scores for techtc500, $MI=0.05$,
    $FE=1000$, $t=0.1$}
\end{figure}

See Figures \ref{fig:fsc3} and \ref{fig:msc3}; one can check
that the average feature selection and learning scores are similar
with and without metalearning.

\begin{figure}
  \includegraphics[scale=0.38]{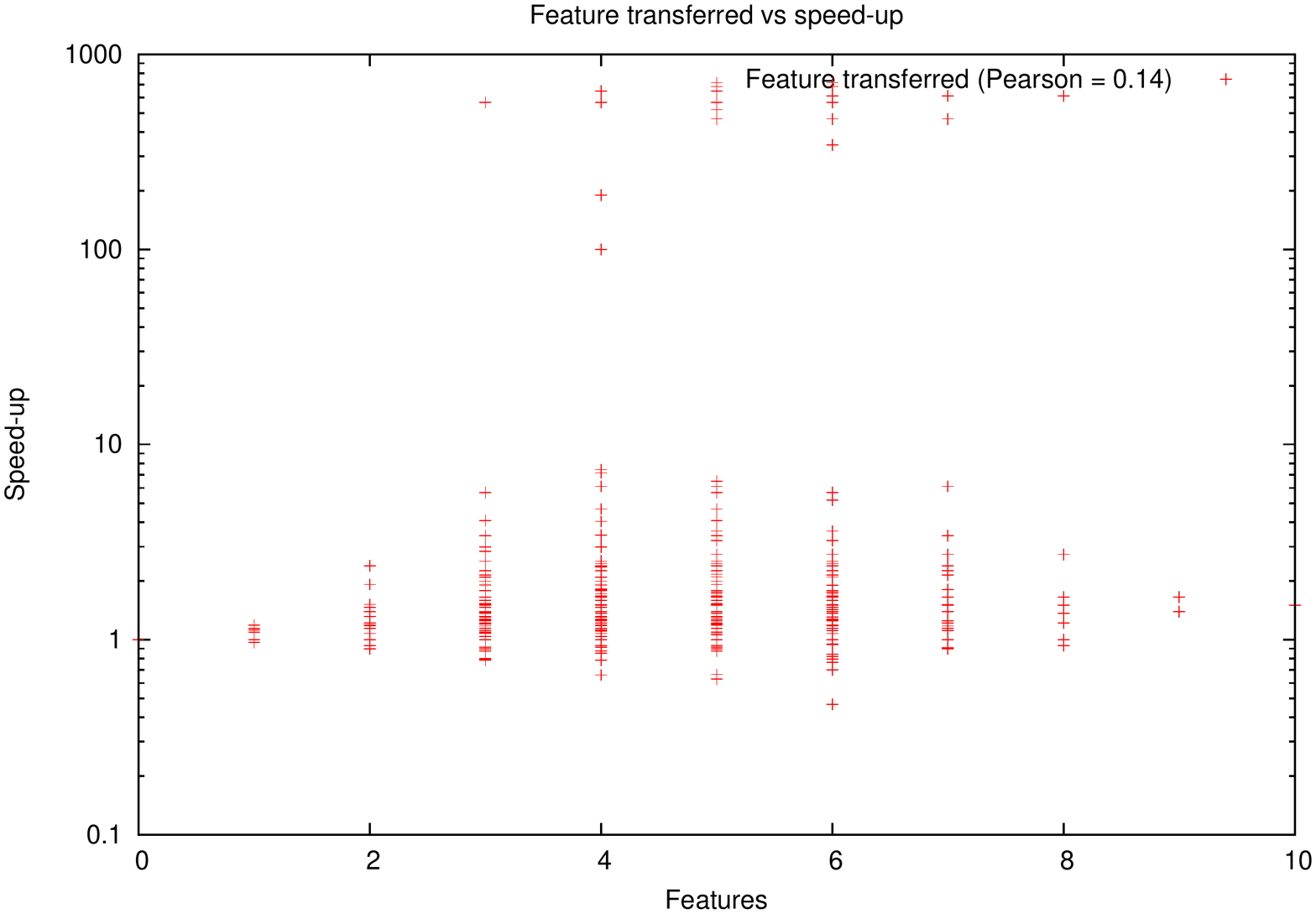}
  \caption{\label{fig:fnsuc1}Feature transferred vs Speed-up for techtc500,
    $MI=0.05$, $FE=1000$, $t=0.1$}
\end{figure}

As it might be expected there is a correlation between the number of
features transferred and speed-up; Figure \ref{fig:fnsuc1} shows a
Pearson correlation of 0.14, not high but substantial.

\begin{figure}
  \includegraphics[scale=0.38]{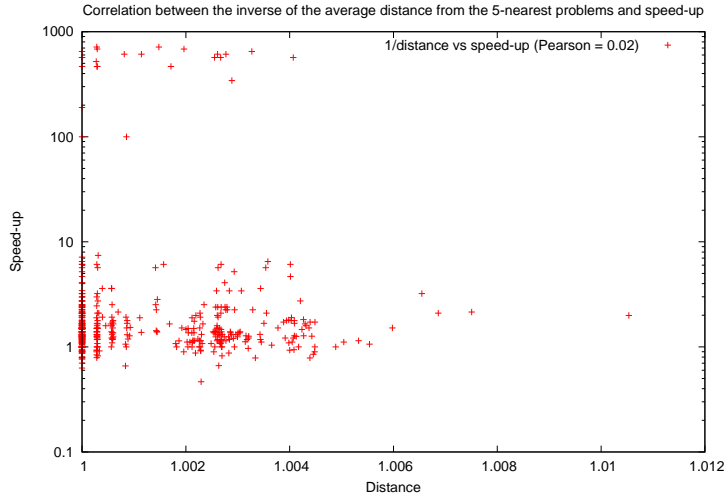}
  \caption{\label{fig:dsuc1}Inverse of distance vs Speed-up for
    techtc500, $MI=0.05$, $FE=1000$, $t=0.1$}
\end{figure}

One may wonder what portion of this correlation is attributable to the
distance. However Figure \ref{fig:dsuc1} shows a quite low Pearson
correlation of 0.02 between the inverse of distance and speed-up. This
low correlation is explained by the fact that the quality of the
feature to be transferred plays an important role in the
heuristic used for feature transfer.

Overall, in the experiments reported just above,
we have observed some impressive speed-up for the techtc500 due to
the greater number of datasets in the collection; and also more
importantly, due to the higher similarity between these datasets, as all web pages
constituting the datasets were drawn from the categories
/Top/Arts/Music and /Top/Science/Math. However the number of
evaluations for feature selection was still rather low in these experiments. Let see now how
it goes when that number is 10000 instead of 1000.

\subsection{High feature selection effort, very high transfer
  threshold}
\label{ref:realdeal}
\begin{figure}
  \includegraphics[scale=0.38]{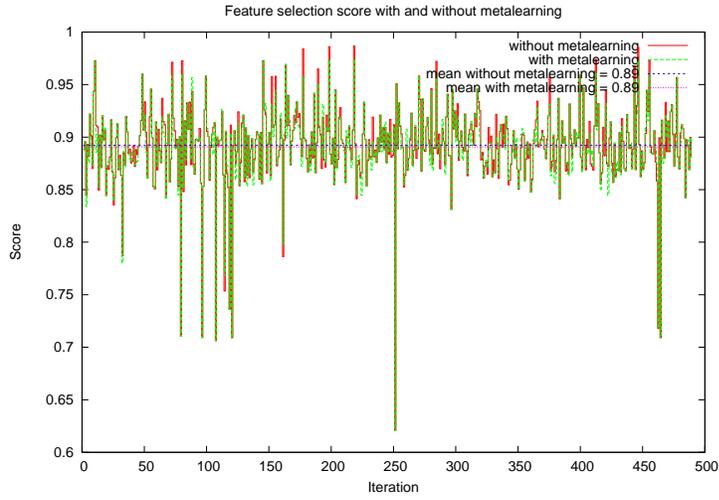}
  \caption{\label{fig:fsc4}Feature selection scores for techtc500,
    $MI=0.05$, $FE=10000$, $t=0.2$}
\end{figure}

\begin{figure}
  \includegraphics[scale=0.38]{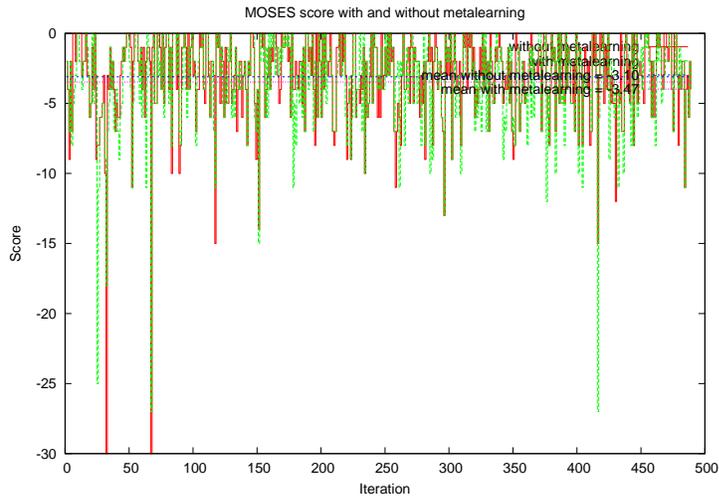}
  \caption{\label{fig:msc4}Learning scores for techtc500, $MI=0.05$,
    $FE=10000$, $t=0.2$}
\end{figure}

This experiment sets the number of evaluation for feature selection to
$FE=10000$ which leads to higher feature selection and learning scores
(as shown in Figures \ref{fig:fsc4} and \ref{fig:msc4}). We did try
with 100000 but the improvement (in term of feature selection and
learning scores) was not really substantial so this setting is ``the
real deal''. Due the highest quality target we need to increase the
transfer threshold to $t=0.2$ to measure positive transfer learning.

\begin{figure}
  \includegraphics[scale=0.38]{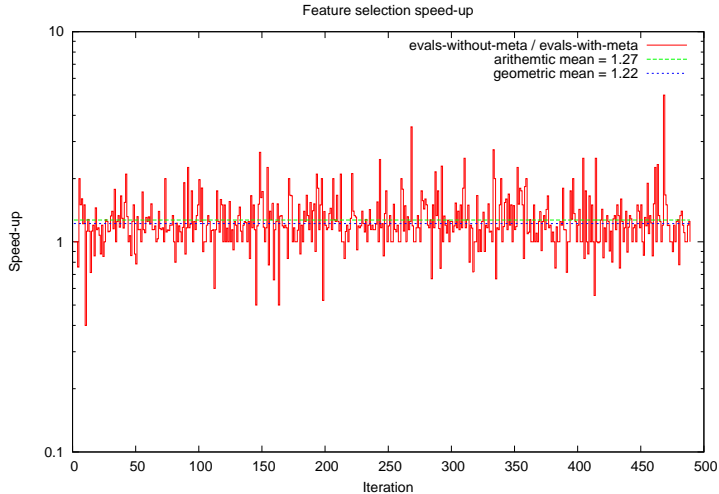}
  \caption{\label{fig:spup7}Speed-up for techtc500, $MI=0.05$,
    $FE=10000$, $t=0.2$}
\end{figure}

As Figure \ref{fig:spup7} shows, although we do not get any massive
speed-ups as in the previous experiment, the speed-up geometric mean,
1.22, is significant and the figure shows again a clear tendency
toward faster feature selection as measure as the experiment
progresses.

Again one can check that the feature selection and learning scores are
roughly the same with and without metalearning. The learning scores
for metalearning are slightly lower while the feature selection scores
are identical. This could be explained because the feature selection
fitness is not in perfect coherence with learning fitness; perhaps
implementing the feature selection confidence as defined in Section
\ref{sec:featconf} would improve that.

\subsection{Metalearning Computational Cost}

We have seen that the feature metalearning methodology is showing
promise even for hard problem collections (hard due to the distant
semantic between problems). However it remains to evaluate whether
feature metalearning is worth it in term of computational cost.  Here
we present conclusions based on extrapolating from the computational cost 
of our experiments, run on an Intel Quad Core
2.5GHz, with feature selection parallelized over the 4 cores. 

First, we measured that the computational cost of MetaDB maintenance
is largely negligible as compared to feature selection -- with the
exception of the nearest neighbor querying process.  Regarding the
latter, we've represented in Figure \ref{fig:time1} the computational
time for nearest neighbor query, beside feature selection with and
without metalearning, for the experiment detailed in Section
\ref{ref:realdeal}, using our current implementation. 

\begin{figure}
  \includegraphics[scale=0.38]{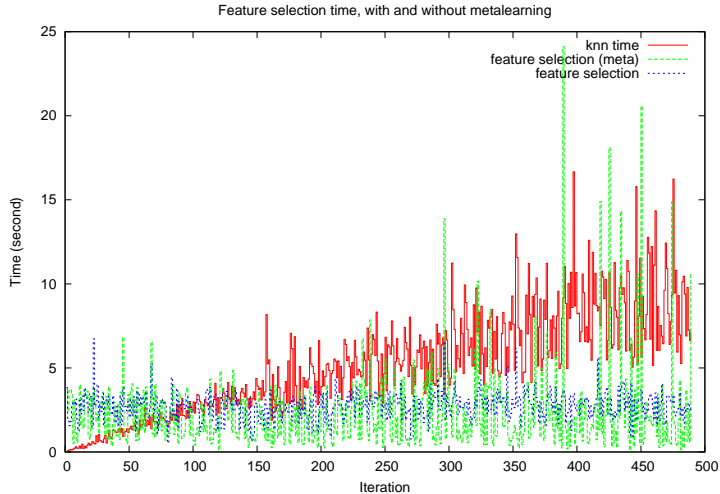}
  \caption{\label{fig:time1}Computational time of metalearning for techtc500, $MI=0.05$,
    $FE=10000$, $t=0.2$}
\end{figure}

As shown in the figure, in our current implementation,
the time spent for nearest neighbor query is
not only significant but badly exceeds the time for feature
selection.  What this shows is that, in our current software implementation,
feature metalearning is not practically useful.   Fortunately, though, this
is a function of the limitations of our current implementation, rather than
reflecting a fundamental limitation of the algorithmic approach.

It is clear that the nearest neighbor query implementation in our current
software is very far from optimal, as it was created solely to allow rapid
experimentation with the feature metalearning methodology.   For one thing,
this part of our current
feature metalearning pipeline has been entirely
coded in Python, while feature selection is coded in C++ and the
computation is spread over 4 cores.  Considering that Python is at least 10x slower than C++ and that
nearest neighbor query is running on a single core while feature
selection is running on 4 cores the results shown in the figure are
quite unfair.  The following Figure shows the same graph but with nearest neighbor
query time divided by $10\times 4 = 40$.

\begin{figure}
  \includegraphics[scale=0.38]{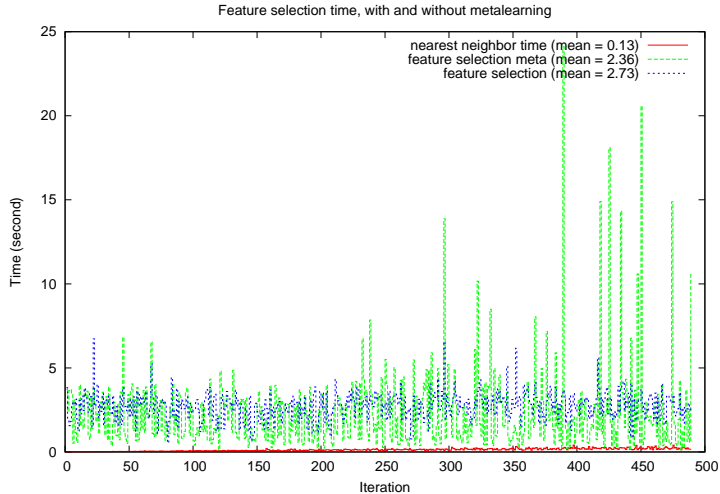}
  \caption{\label{fig:time1}Computational time of metalearning for techtc500, $MI=0.05$,
    $FE=10000$, $t=0.2$}
\end{figure}

Here the estimated cost of metalearning maintenance is almost
negligible but shows a linear increase which -- if it continued -- would eventually render
the approach too costly.  We conjecture that at some point the cover tree data structure
used in the MetaDB
will dampen the linear growth before the cost becomes intolerable.
There are also other ways to address this problems, for instance one could
consider datasets containing only the most relevant features instead
of the entire dataset when initially calculating the JSD;
and then only calculate the JSD on the entire datasets if the distance estimated over
the filtered datasets goes under a certain threshold.

\section{Conclusion and Future Directions}

The results of our initial explorations in feature metalearning,
presented here, are very promising in the sense that they demonstrate that the
approach works.  However, a lot of work remains to turn feature
metalearning into a broadly powerful methodology.

Specifically, we have shown that, in the text classification context at least,

\begin{itemize}
\item If the collection of datasets is sufficiently "dense", in the sense that 
a dataset generally has some fairly close neighbors in dataset-space, then
feature metalearning gives positive speedup and doesn't degrade the quality
of the final answer
\item The speedup is greatest in the case where one needs a quick answer
and doesn't allocate enough time to the problem-solving process to find an
optimal feature set, but only needs a "best I can find in the time available"
feature set
\end{itemize}

In order to work around the first limitation, we believe it will be
necessary to replace nearest-neighbor search with a more sophisticated
meta-level learning algorithm -- perhaps, for instance, using
evolutionary programming itself on the meta level, to learn patterns
in which feature combinations tend to be effective on which problems.

In order to work around the second limitation, and perhaps to an extent the
first as well, the following steps may be useful:

\begin{enumerate}
\item Improve the feature transfer formula. The one given in Section
  \ref{sec:heu} is a rather crude heuristic; given more work we could
  improve its positive vs negative transfer ratio.
\item The more features allowed to be selected, the more likely
  positive transfer may occur. Here, again because running such large
  experiments takes a lot of time, the experiments were conducted with
  only a dozen features on average for each iteration. The typical
  number of features required to solve the family of problems we
  explored are more commonly around 50 and can go into hundreds.
  \end{enumerate}
  
It should also be noted that the problem collection we chose to work
  with is quite hard from a feature metalearning perspective, because the problems are semantically
  quite distant.  For this among other obvious reasons, it would be nice
  to explore feature metalearning on a variety of different problem areas,
  not just supervised text classification of Web pages.

We believe the feature metalearning approach has great promise to dramatically
accelerate machine learning in cases where there is a large number of relatively similar problems
with features drawn from a common feature space.  The initial algorithms reported here
have sufficed to demonstrate the viability of the approach, and explore some of its properties.
Refining them via future research should ultimately lead to the development of extremely
powerful feature metalearning systems.

\paragraph{Acknowledgement} We would like to thank Andras Kornai for encouraging us to look
at the problem of feature metalearning.

\bibliographystyle{alpha}
\bibliography{ML_technical_report}

\begin{thebibliography}{DWBB09}

\bibitem[BCSV08]{Brazdil}
Pavel Brazdil, Christophe~Giraud Carrier, Carlos Soares, and Ricardo Vilalta.
\newblock {\em Metalearning: Applications to Data Mining}.
\newblock Springer, 2008.

\bibitem[BKL06]{Beygelzimer06covertrees}
Alina Beygelzimer, Sham Kakade, and John Langford.
\newblock Cover trees for nearest neighbor.
\newblock In {\em ICML}, pages 97--104, 2006.

\bibitem[Bla09]{Blachnik}
M~Blachnik.
\newblock Comparison of various feature selection methods in application to
  prototype best rules advanced in intelligent and soft computing.
\newblock In {\em Proc. of 6th Inter. Conf. on Comp. Recognition Systems,
  CORES'09}, 2009.

\bibitem[DGM04]{DGM04}
Dmitry Davidov, Evgeniy Gabrilovich, and Shaul Markovitch.
\newblock Parameterized generation of labeled datasets for text categorization
  based on a hierarchical directory.
\newblock In {\em The 27th Annual International ACM SIGIR Conference,
  Sheffield, UK, July}, pages 250--257, 2004.

\bibitem[DWBB09]{Duch}
W~Duch, T~Wieczorek, J~Biesiada, and M~Blachnik.
\newblock Comparision of feature ranking methods based on information entropy.
\newblock In {\em Proc. of 6th Inter. Conf. on Comp. Recognition Systems,
  CORES'09}, 2009.

\bibitem[Gab04]{techtc300}
Evgeniy Gabrilovich.
\newblock http://techtc.cs.technion.ac.il/techtc300/techtc300.html, 2004.

\bibitem[Gei11]{techtcbuilder}
Nil Geisweiller.
\newblock https://github.com/ngeiswei/techtc-builder, 2011.

\bibitem[GIGH08]{PLN}
B~Goertzel., M~Ikle, I~Goertzel, and A~Heljakka.
\newblock {\em Probabilistic Logic Networks}.
\newblock Springer, 2008.

\bibitem[GSMP08]{Biomind}
Ben Goertzel, Lucio Souza, Mauricio Mudado, and Cassio Pennachin.
\newblock Identifying the genes and genetic interrelationships underlying the
  impact of calorie restriction on maximum lifespan: An artificial intelligence
  based approach.
\newblock {\em Rejuvenation Research}, 2008.

\bibitem[KDBB10]{Kachel}
A~Kachel, W~Duch, Blachnik, and M~Biesiada.
\newblock Infosel++: Information based feature selection c++ library.
\newblock In {\em ICAISC 2010,Lecture Notes in Artificial Intelligence, Vol.
  6113}, pages 388--396, 2010.

\bibitem[Loo06]{Looks2006}
Moshe Looks.
\newblock {\em Competent Program Evolution}.
\newblock PhD Thesis, Computer Science Department, Washington University, 2006.

\bibitem[Loo07]{Scalable}
M.~Looks.
\newblock Scalable estimation-of-distribution program evolution.
\newblock In {\em Genetic and evolutionary computation conference}, 2007.

\bibitem[Net]{DMOZ}
Netscape.
\newblock http://www.dmoz.org.

\end{thebibliography}
\end{document}